\documentclass[runningheads]{llncs}
\usepackage[T1]{fontenc}
\usepackage{graphicx}
\usepackage{booktabs}
\usepackage{multirow}
\usepackage{tabularray}
\usepackage{tabularx}
\usepackage{amsmath}
\usepackage{amsfonts}
\usepackage{amssymb}
\usepackage{mhchem}
\usepackage{subfigure}
\usepackage{stmaryrd}
\usepackage[pagebackref=true,breaklinks=true,colorlinks=true,bookmarks=false,citecolor=blue,urlcolor=blue,linkcolor=blue]{hyperref}
\usepackage{booktabs}
\usepackage{makecell}
\usepackage{adjustbox}
\usepackage{pifont}
\usepackage{marvosym}
\begin{document}
\title{FedIA: Federated Medical Image Segmentation with Heterogeneous Annotation Completeness}
\titlerunning{Federated Learning with Heterogeneous Annotation Completeness}
%
%
\author{Yangyang Xiang\thanks{Equal contribution.}\inst{1} 
\and
Nannan Wu\inst{\star 1}
\and
Li Yu\inst{1} 
\and
Xin Yang\inst{1} 
\and 
\\Kwang-Ting Cheng\inst{2} 
\and
Zengqiang Yan\textsuperscript{1(\Letter)}
}
\authorrunning{Y. Xiang et al.}
%
\institute{School of Electronic Information and Communications, Huazhong University of Science and Technology\\
 \email{\{xyy\_2001,wnn2000,hustlyu,xinyang2014,z\_yan\}@hust.edu.cn}\\
 \and School of Engineering, Hong Kong University of Science and Technology 
        \email{timcheng@ust.hk}
}
\maketitle              
\setcounter{footnote}{0}
\begin{abstract}
    
    Federated learning has emerged as a compelling paradigm for medical image segmentation, particularly in light of increasing privacy concerns. However, most of the existing research relies on relatively stringent assumptions regarding the uniformity and completeness of annotations across clients. Contrary to this, this paper highlights a prevalent challenge in medical practice: incomplete annotations. Such annotations can introduce incorrectly labeled pixels, potentially undermining the performance of neural networks in supervised learning. To tackle this issue, we introduce a novel solution, named FedIA. Our insight is to conceptualize incomplete annotations as noisy data (\textit{i.e.}, low-quality data), with a focus on mitigating their adverse effects. We begin by evaluating the completeness of annotations at the client level using a designed indicator. Subsequently, we enhance the influence of clients with more comprehensive annotations and implement corrections for incomplete ones, thereby ensuring that models are trained on accurate data. Our method's effectiveness is validated through its superior performance on two extensively used medical image segmentation datasets, outperforming existing solutions. The code is available at \url{https://github.com/HUSTxyy/FedIA}.

\end{abstract}
\section{Introduction}

Recent progress in federated learning (FL) \cite{FedAvg} has facilitated the collaborative training of unified models across multiple decentralized entities in a privacy-preserving manner \cite{dou2021federated,FPL_CVPR23,FedISM,wu2023federated}. In medical domains, FL has seen extensive application in training segmentation models for distinct lesions and organs \cite{wang2022personalizing,jiang2023fair,wu2023feda3i}. Nevertheless, an essential limitation in current research is the insufficient consideration of the diversity in \textit{\textbf{annotation completeness}} among clients.

\begin{figure}[!t]
    \centering
    \includegraphics[width=0.7\textwidth]{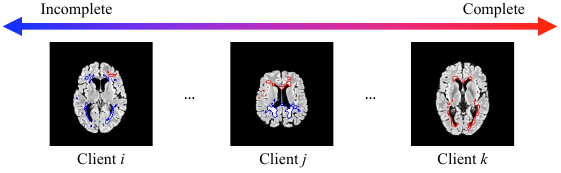}
    \caption{Heterogeneity in annotation completeness among clients. Red and blue solid lines represent the boundaries of marked lesions and unmarked lesions, respectively.}
    \label{fig:bg}
\end{figure}

This issue primarily stems from the varying standards of annotation adopted by various clients. As depicted in Fig. \ref{fig:bg}, certain clients (\textit{i.e.}, client \textit{k}) provide complete annotations for comprehensive diagnosis and analysis. Conversely, others (\textit{i.e.}, client \textit{i} and \textit{j}) may possess incomplete annotations where only partial regions are marked, to minimize labeling costs, which are adequate only for basic image-level assessments (\textit{e.g.}, rapid screening).

Given this heterogeneity in annotation completeness, training a segmentation model via FL poses significant challenges. The inclusion of clients with incomplete annotations creates a situation where these clients are considered to be of lower quality since partial positive regions are mislabeled as background. Such imperfect annotations can negatively affect the overall performance of the model due to the memory effect of neural networks \cite{ELR,ADELE}. To tackle this, in this paper, we focus on the important yet under-explored problem: \textit{\textbf{How to pursue better FL under heterogeneity in annotation completeness?}}

Within the realm of FL, there has been some work focusing on data heterogeneity~\cite{FCCL_CVPR22,FCCLPlus_TPAMI23,FLSurvey}, but the heterogeneity in annotation completeness has been often overlooked. As for strategies to diminish the negative impact of clients with low-quality labels, these solutions predominantly focus on the classification task \cite{FedCorr,FedNoRo}, which is suboptimal when applied to the segmentation task. Although FedA$^3$I \cite{wu2023feda3i} has recently addressed the heterogeneity in annotation quality specific to segmentation, its underlying assumption, where mislabeled pixels mainly distribute near objects' boundaries, renders it ineffective against the challenge of incomplete annotations. Consequently, developing an effective approach to address this critical issue remains an area in need of further exploration and insight.

In this study, we tackle the pressing problem of heterogeneity in annotation completeness by introducing \textbf{FedIA}, a FL solution that is cognizant of and adaptively corrects for client annotation completeness. Our foundational insight is to perceive incomplete annotations as akin to noisy data. We commence by developing an early model robust against the noise associated with incomplete annotations, which then serves as a basis for evaluating each client's level of annotation completeness. Subsequently, our aggregation process prioritizes clients with higher annotation completeness, and clients undertake annotation corrections before local model updating supervised by incomplete annotations. Our approach has been tested on two real-world medical datasets: a brain multiple sclerosis MRI dataset and a COVID-19 lesion CT dataset. Results show that FedIA outperforms other SOTA methods designed to address noisy/imperfect annotations.

The contributions of this paper are three-fold: (1) A new FL problem concentrating on heterogeneity in annotation completeness; (2) A novel solution named FedIA to tackle incomplete annotations; (3) Extensive evaluation to demonstrate the superiority of the proposed solution.

\section{Methodology}
\subsection{Preliminaries and Overview}
This paper focus on a a single-class multi-lesion\footnote{This means an image can contain multiple lesions, each forming a connected region.} segmentation problem in a federated scenario. Given $K$ clients, each client possesses its private dataset 
$D_k=\{\left(x_i \in \mathcal{X} \subseteq \notag \right. \left.
\mathbb{R}^{\mathrm{H} \times \mathrm{W} \times \mathrm{C}}, \tilde{y}_i \in \mathcal{Y}=\{0,1\}^{\mathrm{H} \times \mathrm{W}}\right)\}_{i=1}^{n_k}$,
where ${n_k}$ is the size of $D_k$ and $\left(x_i,\tilde{y}_i \right)$ represents the image-annotation pair characterized by dimensions: height $\mathrm{H}$, width $\mathrm{W}$, and channel $\mathrm{C}$.
Contrary to an ideal situation, the annotations in our case are considered imperfect due to incompleteness, with not every lesion being marked. The completeness ratio $a_k=c^n_{k,i}/c^g_{k,i}$, indicating the proportion of marked lesions to the total actual lesions within $D_k$, which remains identical among samples in $D_k$ but differs across clients.

\begin{figure}[!t]
    \centering
    \includegraphics[width=1\textwidth]{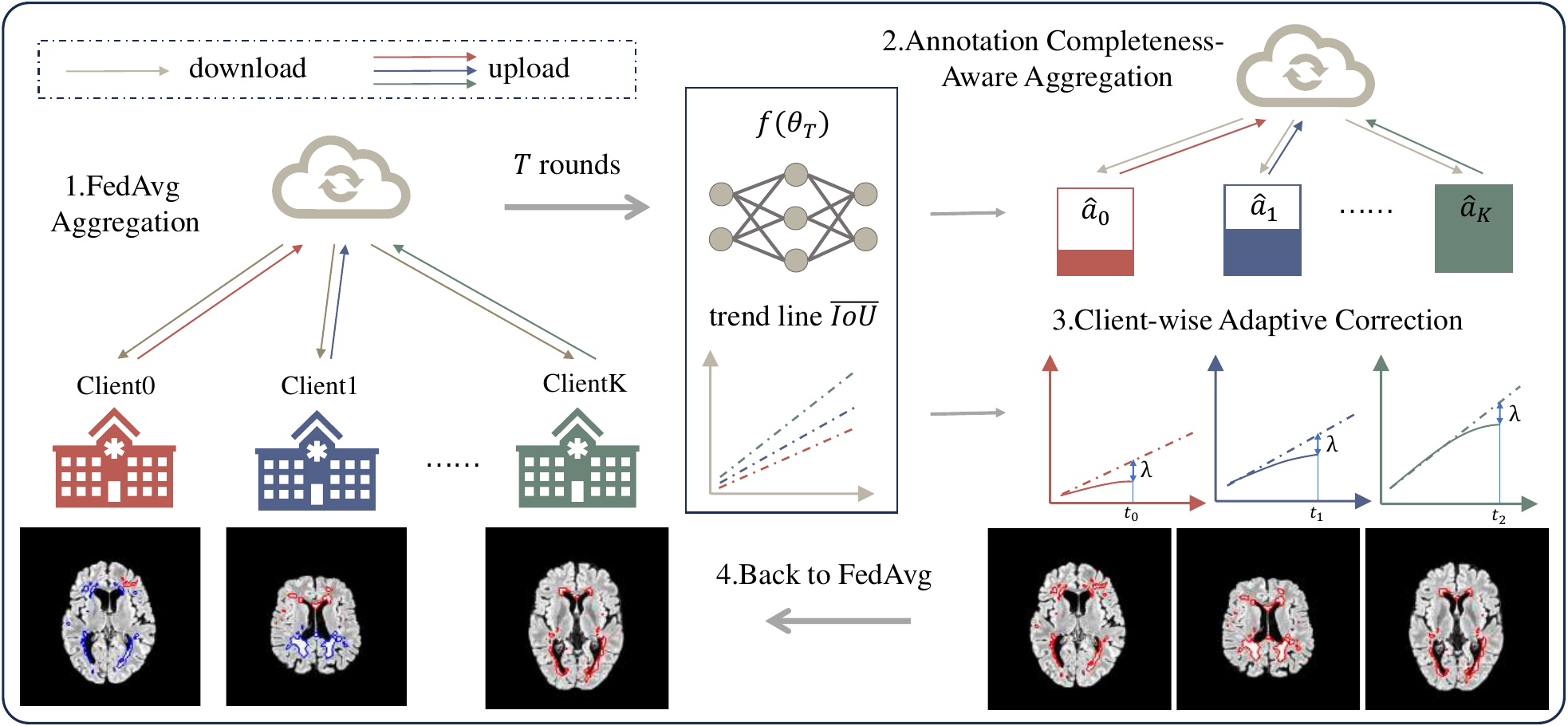}
    \caption{Overview of the proposed FedIA. The first stage is the early learning phase, the global model is updated by FedAvg~\cite{FedAvg}. The second is the modification stage, re-weighting each client by calculating its annotation completeness rate and correcting incomplete annotations synchronously. In the last stage, local models are trained with the corrected labels and aggregated for federated updating through FedAvg~\cite{FedAvg}.}
    \label{method}
\end{figure}

Our objective is to devise a robust algorithm capable of diminishing the negative effects of incomplete annotations on the global model's accuracy. The cornerstone of our approach involves deriving an initial model that is minimally affected by noise through the utilization of extensive noisy data, followed by assessing each client's annotation completeness ratio based on this initial model. The strategy prioritizes learning from clients with higher completeness rates (\textit{i.e.}, higher-quality data), thereby enhancing the model's performance. Furthermore, a mechanism is incorporated to correct incomplete annotations at a certain stage of the learning process, using a specially designed metric based on Intersection over Union (IoU). The overview of FedIA is illustrated in Fig. \ref{method}.

\subsection{Annotation Completeness Estimation}

Assessing the level of annotation completeness across clients is imperative, as it directly influences the tailored handling of each client's data. To accomplish this, obtaining a model that is unaffected by imperfect annotations becomes essential. Our approach begins by interpreting incomplete annotations as noisy labels, with unmarked lesions contributing noise by altering pixel-level labels from 1 to 0. Drawing inspiration from the early learning phenomenon in noisy label learning \cite{ADELE,DivideMix,ELR}, which posits that neural networks initially adapt to clean labels in the early stages before progressively accommodating noisy labels, we can place confidence in the training process despite the prevalence of noisy labels and utilize the early model phase to gauge annotation completeness across clients. Specifically, we develop an early-stage global model parameterized by $\theta_T$, capable of basic segmentation, by undergoing a warm-up period for $T$ communication rounds employing FedAvg \cite{FedAvg}. This process involves the server aggregating client models based on their respective data contributions to formulate the global model, defined as:
\begin{equation}
    \theta_{t}=\sum_{k=1}^K \frac{n_k}{n} \theta_{t,k},
\end{equation}
where $t$ denotes the current training round under the constraint $1 \leq t \leq T$, and $n$ represents the amount of data from all clients, \textit{i.e.}, $n=\sum_{k=1}^K n_k$. In this stage, local optimization of each client is established on:

\begin{equation}
\min_{\theta_k} \sum_{(x_i, y_i) \in D_k} \ell_{d c}\left(f\left(x_i ; \theta_{t,k}\right), \tilde{y}_i\right),
\end{equation}
where $f:\mathcal{X} \rightarrow [0,1]^{\mathrm{H} \times \mathrm{W}}$ is the neural network and $\ell_{d c}:[0,1]^{\mathrm{H} \times \mathrm{W}} \times \mathcal{Y} \rightarrow \mathbb{R}_+$ is the Dice loss~\cite{VNet}.
After obtaining the warm-up model parameterized by $\theta_T$ that is relatively unaffected by noise, the annotation completeness rate $\hat{a}_k$ of each client $k$ is estimated by the following formula:
\begin{equation}
\hat{a}_k=\frac{\sum_{i=1}^{n_k} c_{k,i}^n}{\sum_{i=1}^{n_k} c_{k,i}^p},
\end{equation}%
where $c_{k,i}^n$ and $c_{k,i}^p$ denote the number of lesions in the noisy label $\tilde{y}_i$ and in the predicted map $y_i^p = f(x_i;\theta_T)$, respectively.

\subsection{Annotation Completeness-Aware Aggregation}
Quantity-based aggregation (\textit{i.e.}, FedAvg) is susceptible to noise caused by incomplete annotations, especially when such annotations are numerous \cite{wu2023feda3i}. To mitigate this, clients with higher annotation quality should dominate FL more. Therefore, we calculate a completeness-aware aggregation weight $w_k^t$ at round $t$ for each client $k$, defined as
\begin{equation}
w_k^t=\frac{\exp \left(\frac{\hat{a}_k}{\ell_{k}^t}\right)}{\sum_{j=1}^K \exp \left(\frac{\hat{a}_j}{\ell_{j}^t}\right)},
\end{equation}%
where $\ell_{k}^t$ denotes the average loss of client $k$ at round $t$ calculated by
\begin{equation}
    \ell_{k}^t=\frac{\sum_{(x_i, y_i) \in D_k} \ell_{d c}\left(f\left(x_i ; \theta_{t,k}\right), \tilde{y}_i\right)}{n_k}.
\end{equation}
Generally, the observed loss is lower when annotation completeness is elevated. Consequently, the server prioritizes clients exhibiting lower losses, effectively reducing the negative effects of imprecise estimation of $a_k$ on the weighting process, potentially arising from inappropriate selection of $T$.

\subsection{Client-wise Adaptive Correction}
The volume of data significantly influences the performance of neural networks, and datasets characterized by low annotation completeness represent valuable resources that should not be overlooked. Hence, rectifying incomplete annotations to acquire cleaner data for further training of the model is essential. Given that different clients pose datasets with varying levels of annotation completeness, the onset of noise impact and their robustness to noise vary accordingly.

To capture this information, in the early learning phase (\textit{i.e.}, $1 \leq t \leq T$), we compute IoU values every round for each client $k$ and fit it with the first-order polynomial function:
\begin{equation}
\overline{\mathrm{IoU}}_k(t)=l_k \cdot t+b_k,
\end{equation}
where $l_k$ and $b_k$ are two parameters of the polynomial function. After early learning and in sync with the completeness-aware aggregation, we monitor the change of $\mathrm{IoU}_k$ every round. Any client satisfying the following formula will correct its annotations in the next round:
\begin{equation}
\overline{\mathrm{IoU}}_k(t)-\mathrm{IoU}_k(t)>\lambda,
\end{equation}
where $\mathrm{IoU}_k(t)$ denotes the actual $\mathrm{IoU}$ value of client $k$ at round $t$. $\lambda$ is an adjustable hyperparameter, set to 0.03 by default. In addition, the client only corrects annotations for which its model output predicted probability has confidence above a certain threshold setting of 0.8. It is worth noting that we only correct the pixels with a value of 0 because only false negative lesions and no false positive lesions are presented in this setting.

\section{Experiments}
\subsection{Datasets and Implementation Details}
\subsubsection{Datasets.} Two public medical image segmentation datasets are included:
\begin{enumerate}
    \item Two real-world multiple sclerosis datasets, focusing on the segmentation of white matter lesions (WML) in 3D magnetic resonance (MR) brain images, denoted as \textbf{MS}, including MSSEG-1~\cite{MSSEG} and PubMRI~\cite{PubMRI}. In the task, we only use the FLAIR modality, in which the lesions are relatively clear.
    \item The widely-used COVID dataset, aiming at segmentation and quantification of lung lesions caused by SARS-CoV-2 infection from computed tomography (CT) images, denoted as \textbf{LUNG}.\cite{COVID}
\end{enumerate}
Each dataset is divided into training and test sets by a ratio of 8:2, whose training set is then randomly split into four clients. For computational efficiency, all 3D samples are converted into 2D slices and resized to 256$\times$256 pixels. 

To verify the robustness to different degrees of incompleteness of our method, several settings are used for evaluation. Specifically, for \textbf{MS}, the annotation completeness rate of the $k$-th client is set as $20\% \times k-10\% \times m+40\%$. And we conduct four sets of experiments, \textit{i.e.}, $m=0, 1, 2, 3$. For \textbf{LUNG}, three different settings are used, and the completeness rates are formulated as $10\% \times k-30\% \times m+70\%$, where $m=0, 1, 2$.

\subsubsection{Incomplete Annotation Generation}
When doctors or other professionals label multi-lesion data, they tend to label one lesion at the 3D volume level before annotating another. Therefore, to simulate real noise and generate incomplete annotation $\tilde{v}_j$, lesions are randomly removed at the 3D level. This process allows us to mimic real-world conditions more accurately. Specifically, we first set the annotation completeness of each client as $a_k$, which is unknown during training. Then, we calculate the number of lesion-connected components $c_j^n$ in each 3D sample $v_j$ and randomly choose $c_j^n$ lesion regions, where $c_j^n=c_j^g \cdot a_k$. Only the chosen lesion regions are kept as well-annotated while others are set as background (\textit{i.e.}, incomplete/missing annotation).

\subsubsection{Implementation Details.}
In this work, U-Net~\cite{UNet} is adopted as the foundational model architecture for FL. The FL training process is designed to include a total of 300 communication rounds, with each local training phase consisting of a single local epoch. During local training, the model undergoes optimization via the Adam optimizer with momentum terms set to $(0.9, 0.99)$, a batch size of 4, and an initial learning rate of 1e-4. To accommodate the early learning strategy, the initial learning round, $T$, is set as 10 for \textbf{MS} and 40 for \textbf{LUNG}.

\subsection{Comparison with State-of-the-art Methods}
In our analysis, we benchmark FedIA against recent leading methods tailored to address label noise in both classification and segmentation tasks, including ELR (NeurIPS'20) \cite{ELR} and ADELE (CVPR'22) \cite{ADELE}, which leverage the early learning phenomenon to prevent model overfitting to noisy labels; RMD (TMI'23) \cite{RMD}, which mitigates annotation noise in medical imaging through mutual distillation; NR-Dice (TMI'20) \cite{NRdice}, introducing a noise-robust Dice loss to combat noisy labels; FedNoRo (IJCAI'23) \cite{FedNoRo}, designed to manage class-imbalanced noisy data; and FedCorr (CVPR'22) \cite{FedCorr}, employing the LID score to identity noisy clients. Additionally, we incorporate the universally recognized FL framework, FedAvg~\cite{FedAvg}, as a baseline for comparison. \textit{Detailed implementations of these methods are available in the supplementary material.}

Quantitative comparison results on \textbf{MS} and \textbf{LUNG} measured by Dice coefficient (\%) are summarized in Table \ref{CompareResult}. Notably, our FedIA exhibits consistent performance even as annotation completeness diminishes, in contrast to some methods whose effectiveness wanes with decreased annotation completeness. This demonstrates that FedIA surpasses other sophisticated methods across various datasets and configurations, underscoring the robustness and efficacy of our strategy in tackling the challenge.

 \begin{table}[!t]
    \setlength{\tabcolsep}{3pt} %
    \centering
    \caption{Comparison results with state-of-the-art methods under the \textbf{MS} and \textbf{LUNG} settings. $c_0/c_1/c_2/c_3$ means the annotation completeness rates $a_k$ of clients are $c_k\times10\%$, corresponding to the setting of $m$. The average results (\%) from the last ten rounds are reported. The best results are marked in bold. 
}\label{compare}
    \resizebox{\linewidth}{!}{
        \begin{tabular}{l|l|cccc|ccc}
            \toprule
            \hline
            \multirow{3}{*}{Methods} & \multirow{3}{*}{From} & \multicolumn{4}{c|}{\textbf{MS}} & \multicolumn{3}{c}{\textbf{LUNG}} \\ \cline{3-9} 
            &                                                       &$m=0$             & $m=1$          & $m=2$            & $m=3$       & $m=0$       & $m=1$     & $m=2$\\ 
            &                                                       &4/6/8/10        &3/5/7/9       &2/4/6/8         &1/3/5/7    &7/8/9/10   &4/5/6/7  &1/2/3/4\\ \hline
            FedAvg          & AISTATS'17              & 60.88          & 55.77        & 34.65          & 13.98     &54.40 	 &46.74	   &28.47\\
            ELR                & NeurIPS'20              & 63.01          & 57.97        & 35.91          & 9.50         &49.29	     &35.01	   &11.22\\
            NR-Dice         & TMI'20                  & 69.19          & 64.42        & 60.16           & 23.60      &58.36	     &54.32	   &33.75\\
            ADELE            & CVPR'22                 & 61.34          & 58.63        & 25.01          & 0         &54.33	     &44.14	   &17.89\\
            FedCorr        & CVPR'22                 & 62.03          & 57.12        & 30.35          & 0         &55.24	     &50.56	   &23.78\\
            RMD                & TMI'23                  & 62.77          & 59.15        & 41.60          & 17.15     &48.60	     &32.79	   &9.37\\
		FedNoRo        & IJCAI'23                & 67.09          & 60.78        & 39.74          & 31.53     &49.86	     &37.25	   &19.16\\ \hline
		FedIA                         & Ours                    & \textbf{74.73} & \textbf{74.03} & \textbf{69.22} & \textbf{56.53} &\textbf{59.42}	&\textbf{55.34} &\textbf{44.72}\\ \hline
            \bottomrule
	\end{tabular}
    }\label{CompareResult}
\end{table}

\begin{figure}[!t] 
	\centering
	\subfigure[GT]{\includegraphics[width=0.16\columnwidth]{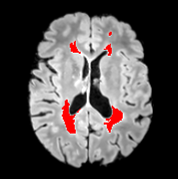}}
	\subfigure[FedIA]{\includegraphics[width=0.16\columnwidth]{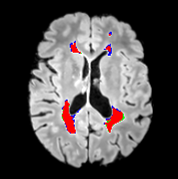}}
	\subfigure[FedNoRo]{\includegraphics[width=0.16\columnwidth]{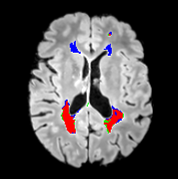}}
	\subfigure[RMD]{\includegraphics[width=0.16\columnwidth]{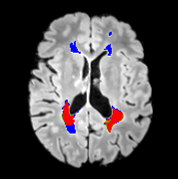}}
	\subfigure[NR-Dice]{\includegraphics[width=0.16\columnwidth]{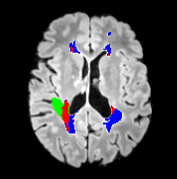}}
	\subfigure[others]{\includegraphics[width=0.16\columnwidth]{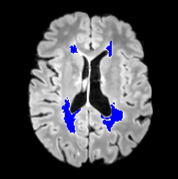}}
	\caption{Qualitative comparison on \textbf{MS} where \textit{others} represents FedAvg, ELR, ADELE, and FedCorr failing to segment any lesion. Red, blue and green color show the prediction of true-positive, false-negative and false-positive regions, respectively.}
	\label{fig:visual}
\end{figure}

Exemplar qualitative comparison on \textbf{MS} in annotation completeness setting of 10\%, 30\%, 50\%, 70\% is illustrated in Fig.~\ref{fig:visual}. FedIA effectively recalls all lesions with fewer false positives, leading to the best segmentation performance. Comparatively, FedNoRo, RMD, and NR-Dice suffer from extensive false negatives, resulting in noisy segmentation. What's worse, FedAvg and other comparison methods completely fail to segment any lesion, indicating the difficulty in addressing heterogeneous annotation completeness and the necessity and value of FedIA. \textit{More qualitative results are available in the supplementary material.}

\subsection{Ablation Study}

\subsubsection{Component-wise Study.}
We conduct an ablation study by separately removing the Annotation Completeness-Aware Aggregation (ACAG) and the Client-wise Adaptive Correction (CAC) components from FedIA as summarized in Table \ref{component}. We observe that FedAvg can benefit from both components, particularly under the lowest annotation completeness settings. This phenomenon demonstrates the effectiveness of our designs against annotation noise. The best performance is typically achieved when both components are incorporated.

\subsubsection{Impact of the Early Learning Round $T$.}
It is worth noting that the early learning phase, denoted by $T$, is set differently for the two tasks as \textbf{LUNG} presenting a more complex learning challenge compared to \textbf{MS}, essentially requiring a longer early learning period. This variation prompts a relevant question regarding the optimal number of training rounds necessary for effective early training. To address this, we perform ablation studies on \textbf{LUNG} under various $T$ settings: 10, 20, 30, and 40 as summarized in Table \ref{warm}. The results indicate that our method exhibits considerable robustness to changes in $T$. Notably, FedIA consistently outperforms the baseline FedAvg across all tested $T$ selections, demonstrating its robustness and superior performance irrespective of the early learning duration. \textit{More ablation studies are available in the supplementary material for reference.}

\begin{table}[!t]\setlength{\tabcolsep}{3pt}
		\centering
		\begin{minipage}{0.58\linewidth}
			\centering
			\caption{Component-wise study.}\label{abalation}
			\begin{tabular}{ccc|cccc}
				\toprule
				\hline
				\multirow{2}{*}{FedAvg} & \multirow{2}{*}{ACAG} &\multirow{2}{*}{CAC}&\multicolumn{4}{c}{\textbf{MS}} \\ \cline{4-7}
    
		&	&&0 & 1        & 2        & 3       \\\hline
				\checkmark     &      &      & 60.88    & 55.77    & 34.65  &13.98 \\
				\checkmark  & \checkmark  &  & 71.91  & 70.34  & 66.95    &43.40 \\
				\checkmark  & & \checkmark  & 74.23  & 73.67  & 67.43  &37.74 \\
				\checkmark & \checkmark & \checkmark &\textbf{74.73} & \textbf{74.03} & \textbf{69.22} & \textbf{56.53}   \\ \hline
				\bottomrule
			\end{tabular}\label{component}
		\end{minipage}
		\begin{minipage}{0.41\linewidth}
			\centering
			\caption{Impact of the Round $T$}\label{param}
			\begin{tabular}{c|ccc}
				\toprule
				\hline
				\multirow{2}{*}{$T$} & \multicolumn{3}{c}{\textbf{LUNG}}  \\\cline{2-4}
    &0 &1 &2 \\ \hline
				10	&56.53 	&52.25	&\textbf{46.88} \\
				20	&59.36 	&54.78	&45.37	\\
				30	&\textbf{59.52}	&55.07	&44.74 \\
				40	&59.42	&\textbf{55.34} &44.72 \\ \hline
				\bottomrule
			\end{tabular}\label{warm}
		\end{minipage}
	\end{table}
 
\section{Conclusion}
In this study, we tackle a significant yet overlooked challenge in federated medical image segmentation: how to enhance FL against heterogeneity in annotation completeness. We approach incomplete annotations as akin to noisy data, employing strategies to mitigate their negative impacts denoted as FedIA. FedIA involves initially assessing the level of annotation completeness at the client level through designed indicators. Then, it prioritizes clients with greater annotation completeness and undertakes corrective measures for those with incomplete ones, aiming to ensure that the training process is mainly based on accurate knowledge. After rigorously evaluated through a line of experiments on two extensively utilized medical image segmentation datasets, experimental results affirm the effectiveness of FedIA, showcasing its advantage over current leading approaches. We believe that the issue formulated and the proposed solution will pave the way for more practical FL applications in complex medical scenarios.
\\
\\
\textbf{Acknowledgement.} This work was supported in part by the National Natural Science Foundation of China under Grants 62202179 and 62271220, and in part by the Natural Science Foundation of Hubei Province of China under Grant 2022CFB585. The computation is supported by the HPC Platform of HUST.

%
%
%
%
\bibliographystyle{splncs04}
\bibliography{reference}

\end{document}


%

%
%
\begin{center}
	\Large
	\textbf{Supplementary Material for}
\end{center}

\begin{center}
	\large
	\textbf{FedIA: Federated Medical Image Segmentation with Heterogeneous Annotation Completeness}
\end{center}

%
\begin{table}[h]%
    \setlength{\tabcolsep}{5pt} %
    \centering
    \caption{Detailed implementations of the other methods. If correct in method, only correct pixel with value 0, the same as FedIA.
}
    \begin{tabular}{c|c}
        \toprule
        \hline
        Methods & Parameters  \\ \hline
         FedAvg &the same as FedIA            \\           
         ELR &  $\beta=0.99,~\lambda=1$         \\
         NR-Dice &$\gamma=1.5$          \\
         ADELE &  $\lambda=1,~\tau=0.8,~\rho=0.8,~r=0.8/0.9~(\textbf{MS}/\textbf{LUNG})$          \\
         FedCorr &$T_1=10,~T_2=140,~T_3=150$           \\
         RMD &$t_1=10,~u=0.8,~\tau=0.005$            \\
         FedNoRo &  T~(warm~up~round) $= 10/20~(\textbf{MS}/\textbf{LUNG})$  \\\hline
        \bottomrule
\end{tabular}
\end{table}


\begin{table}[h]
    \setlength{\tabcolsep}{5pt} %
    \centering
    \caption{Ablation results about the correction threshold $\tau$. We found that a smaller threshold is more appropriate when the annotation completeness is lower, which indicates that the network is not so confident about its prediction when the annotations are incomplete. }
    \begin{tabular}{l|cccc}
        \toprule
        \hline
        \multirow{3}{*}{$~\tau$} & \multicolumn{4}{c}{\textbf{MS}}  \\ \cline{2-5} 
        &$m=0$             & $m=1$          & $m=2$            & $m=3$\\                                                      &4/6/8/10        &3/5/7/9       &2/4/6/8         &1/3/5/7    \\ \hline
        0.5        &74.68	&73.55	&\textbf{71.04}	&\textbf{66.71}\\
        0.6       &74.04	&73.51	&68.06	&65.86 \\
        0.7       & 74.66	&73.13	&69.43	&64.66\\
        0.8       &\textbf{74.73}	&\textbf{74.03}	&69.22	&56.53 
\\ \hline
        \bottomrule
\end{tabular}
\end{table}

\begin{figure}[h]
    \centering
    \includegraphics[width=0.85\textwidth]{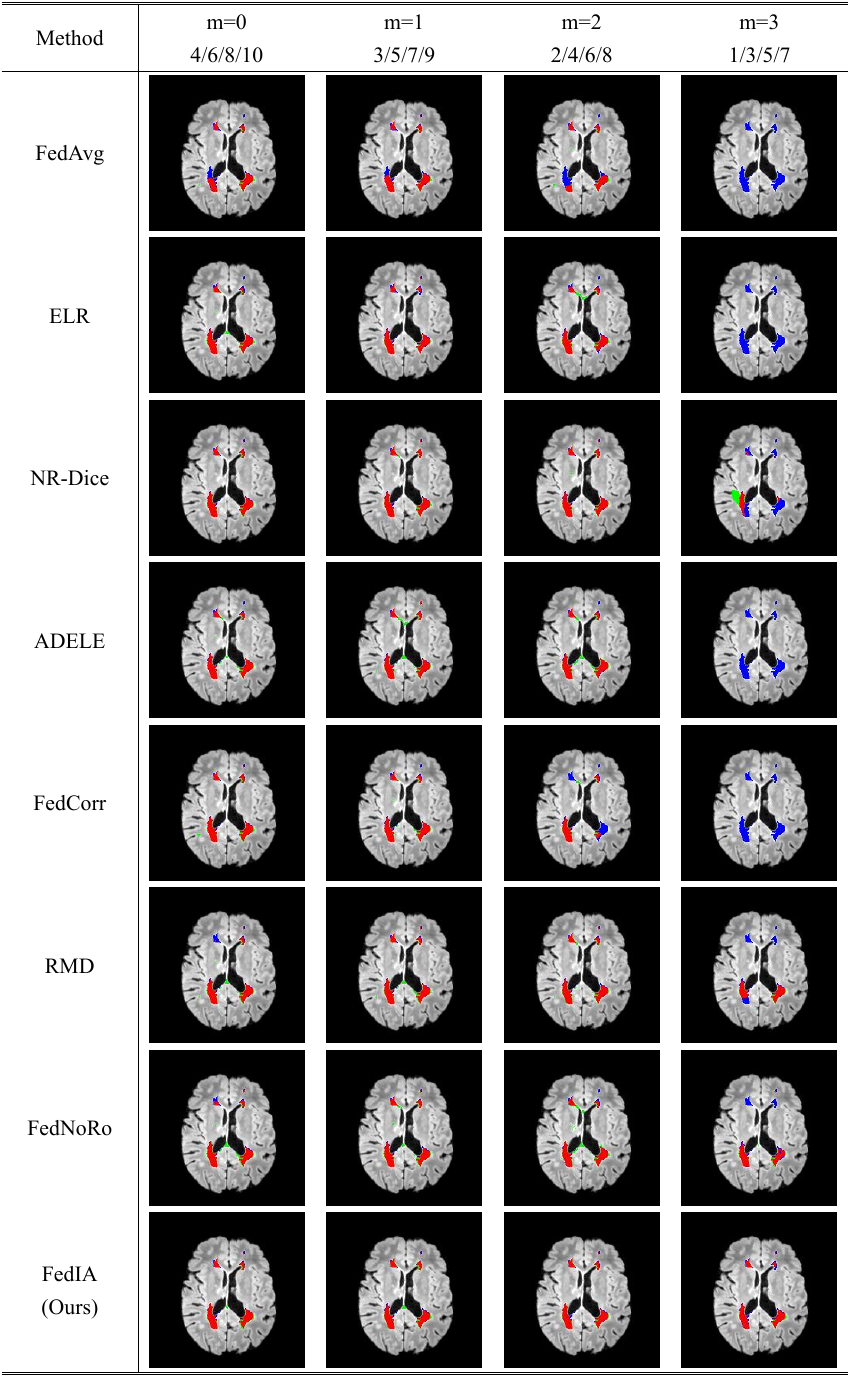}
    \caption{Qualitative comparisons of different methods. We provide qualitative results in different settings on \textbf{MS} for visual comparison. Red, blue and green color show the prediction of true-positive, false-negative and false-positive regions, respectively.}
    \label{fig:vis}
\end{figure}